\documentclass[a4paper, 10pt, conference]{ieeeconf}      

\IEEEoverridecommandlockouts                              

\usepackage{graphicx} 

\usepackage{fancyhdr}
\usepackage{eso-pic}

\title{\LARGE \bf
The Influence of Facial Features on \\the Perceived Trustworthiness of a Social Robot*
}

\author{Benedict Barrow$^{1}$ and Roger K.\ Moore$^{2}$
\thanks{*This study was conducted by B.\ Barrow (supervised by Prof.\ R.\ K.\ Moore) as part of his MComp.\ degree in AI \& Computer Science at the University of Sheffield.}
\thanks{$^{1}$Benedict Barrow has now graduated from the University of Sheffield, UK
        {\tt\small benedict.barrow@gmail.com}}%
\thanks{$^{2}$Roger K. Moore is with the School of Computer Science, University of Sheffield, UK
        {\tt\small r.k.moore@sheffield.ac.uk}}%
}

\begin{document}

\AddToShipoutPictureBG*{%
  \AtPageUpperLeft{%
    \setlength\unitlength{1in}%
    \hspace*{\dimexpr0.5\paperwidth\relax}
    \makebox(0,-0.75)[c]{\small SCRITA and RTSS @ IEEE RO-MAN Workshop on TRUST, 29 August 2025, Eindhoven, Netherlands}%
    }}
  
\maketitle
\thispagestyle{empty}
\pagestyle{empty}

\begin{abstract}

Trust and the perception of trustworthiness play an important role in decision-making and our behaviour towards others, and this is true not only of human-human interactions but also of human-robot interactions. While significant advances have been made in recent years in the field of social robotics, there is still some way to go before we fully understand the factors that influence human trust in robots.  This paper presents the results of a study into the first impressions created by a social robot's facial features, based on the hypothesis that a `babyface' engenders trust.  By manipulating the back-projected face of a Furhat robot, the study confirms that eye shape and size have a significant impact on the perception of trustworthiness.  The work thus contributes to an understanding of the design choices that need to be made when developing social robots so as to optimise the effectiveness of human-robot interaction.

\end{abstract}


\section{Introduction}

Trust is a fundamental building block for any society to function properly. We are constantly asked to trust in our government and officials, our infrastructure and institutions, the people we deal with as we go about our business, as well as the technology on which we have all come to rely in our daily lives. Conversely, we are ourselves trusted to perform tasks, to honour confidences, to protect property, to defend the rights of marginalised groups and to safeguard the vulnerable. Without trust, life would be full of risk and difficulty.

Trust might therefore seem to be of obvious philosophical interest, yet, modern research in this area did not begin in earnest until the 1950s \cite{Deutsch1958}. As a multi-faceted construct, there is no unifying definition or framework within which the concept of trust can be analysed. It is clear from research in the social sciences that trust is rooted in social experience and goes far deeper than mere reliance \cite{Cominelli2021}. A key aspect is the concept of putting oneself in a vulnerable position relative to someone else. This necessitates a reliance on the goodwill and kindness of the interactional partner \cite{Baier2014}.  Hence, trust will always involve a certain degree of risk because of imperfect information.

How then does the notion of trust in human-human relationships manifest itself in human-robot relationships? As robot capabilities grow, we are seeing social robots enter our public spaces, homes and workplaces.  Hence, an exploration of the nature of our desired relationships with these new social entities will help to lay the foundations for fruitful and effective human-robot interaction (HRI) in the future.

To address these issues, this paper presents the results of a study into the influence of facial features on the perceived trustworthiness of a social robot.  Section~\ref{sec:THRI} explores the factors that influence trust in HRI, including the importance of first impressions.  Sections~\ref{sec:AIMS} and \ref{sec:METH} outline the objectives of this study and how they were addressed.  Section~\ref{sec:EXP} describes the experiment, and Section~\ref{sec:RES} presents the main findings.  Finally, Section~\ref{sec:CONC} discusses some of the issues encountered in the study.

\section{Trust in Human-Robot Interaction} \label{sec:THRI}

\subsection{Factors Influencing Trust in HRI}

There are a myriad of factors which influence trust in human-robot interactions. According to the `Human-Robot Trust Model' proposed by Sanders at al. \cite{Sanders2011}, these factors can be divided into three categories: (i) those relating to the individual, (ii) those relating to the external environment, and (iii) those related to the robot itself.

\subsubsection*{Human Factors}
Human expectations about the capabilities of a robot can heavily influence how a robot is perceived. If there is a misalignment between expectations and the actual capabilities of a robot, this can lead to a profound loss of trust, especially when mistakes or misunderstandings occur \cite{Kwon2016}. Also, it has been shown that users with greater robot exposure place more trust in robots than those with less experience \cite{Sanders2017}, and a user's personality (such as introversion/extroversion) also plays a role \cite{Robert2018,Huang2023d}.

\subsubsection*{External Factors}
Trust in HRI can be affected by the nature of the task at hand. For example, in high-risk situations such as rescue missions, trust is a critical factor that influences human decision-making and the attendant outcomes \cite{PARK2008}. Also, cultural factors can condition differences in attitudes towards robots \cite{Wang2010}.

\subsubsection*{Robot Factors}
Since trust develops from a sequence of interactions over time, any unpredictable behaviour or system malfunction undermines confidence in the relationship \cite{Sanders2011}. Furthermore, attributes such as a robot's physical appearance, its voice, its anthropomorphic qualities, its general demeanour and its potential for uncanniness also exert an influence \cite{Goetz2003,Hancock2011,Eyssel2012,Mori1970,Moore2012}.  

\subsection{First Impressions}

First impressions matter, and it has been shown that people make incredibly quick\footnote{Less than 100 msec.} initial judgements about personality traits, including trustworthiness, from the evaluation of facial features \cite{willis2006first}.  This means that, regrdless of whether such judgements are accurate or not, facial stereotypes guide decision-making and interpersonal interactions \cite{van2008friend}.

Of course, such implicit biases can be problematic in specific contexts. For example, it has been demonstrated in court sentencing that facial stereotypes have real-life legal consequences \cite{berry1988s}. In particular, it has been shown that facial maturity was a predictor of the outcomes of court rulings, with `babyfaced' defendants being found guilty less often \cite{zebrowitz1991impact}. These findings were replicated in a 2020 study in which guilty verdicts were 8.03\% higher in defendants who were perceived to be ``untrustworthy-looking'' \cite{jaeger2020can}.

\subsection{Facial Features}

As discussed above, the factors which influence trust in social robots are complex and considerable in number. Robot-related features have been shown to have the most impact and, due to the importance of the face in understanding social cues and intention, it is interesting to consider how facial features might be designed in order to create a trustworthy first impression. 

Altering the physical features of a robot can be difficult and costly. However, new technologies (such as back projection) allow a face to be animated and rendered onto a screen which makes exploring these design features more accessible \cite{kalegina2018characterizing}. This opens up new avenues for the investigation of how facial features such as the shape of the eyes, eyebrows and face might be designed in order to create stronger and more trustworthy first impressions of social robots.

\section{Aims of this Study} \label{sec:AIMS}

The effect of adjusting the size, shapes and positions of facial features have already been explored within the context of social robotics.  For example, in a study by Song et al.\  \cite{song2021effect}, participants were shown potential designs for a social robot face followed by a trustworthiness evaluation. However, the designs were two-dimensional and shown to the participants via a computer screen.  Hence, it is possible that there are differences between the perception of a two-dimensional representation of a social robot versus a three-dimensional, co-located embodied agent.  

Some work has been carried out using an embodied social robot by adjusting facial expressions \cite{Ghazali2018} or social cues \cite{desteno2012detecting} but, to the authors' knowledge, the effect of size adjustment, shape and positioning of facial features on eliciting trust in an embodied social robot has not been carried out hitherto.  Hence the main objective of the study reported here was to address the question -- can the influence of `babyfaced' features on trust be replicated in HRI?

The `baby schema' (German: \emph{Kindchenschema}) is a phenomenon in which certain babyfaced features can positively influence impressions of a person’s character. According to Zebrowitz and Montepare \cite{zebrowitz1992impressions}, babyfaced individuals are characterised by having large eyes, round faces, thin eyebrows and small nose bridges, and the combination of these physical properties are positively correlated with perceived trustworthiness.  Therefore, building on these findings, the study reported here tested the hypothesis that larger eyes and/or rounder eyes and/or thinner eyebrows would increase the initial impression of trustworthiness in a social robot.

\section{Methodology} \label{sec:METH}

\subsection{The Furhat Robot}

Furhat is a social robot developed by Furhat Robotics\footnote{https://www.furhatrobotics.com}. The Furhat robot has a human-like head which uses back projection to display a digital face on a translucent, physical mask in real-time. This means that the face can be highly customised, animated and adapted on-the-fly. A variety of physical masks are available, allowing for wide variation in the size and shape of projected faces.

Furhat ships with a number of pre-configured faces which can be interchanged very easily. It is also possible to create custom faces or manipulate these presets, which are essentially textures mapped onto a a 3D model and then projected onto a mask.

While these textures can be manually adjusted in photo-editing software, it is also possible to manipulate facial features programmatically. For example, it is possible to adjust values of variables such as \texttt{MOUTH\_NARROWER} or \texttt{EYES\_WIDER}.  Hence this flexibility makes Furhat an ideal platform for investigating the hypothesis mentioned above.

\subsection{The Trust Game}

Within the field of social robotics there is a great deal of variety in how trust is measured. Self-report measures are common, which often employ non-standardised custom scales, or behavioural tasks, such as game-playing \cite{chita2021can}.  

One example of such a behavioural measure is borrowed from game theory. The `Trust Game' \cite{berg1995trust} is an economic game which models trust as the amount of financial risk a participant is willing to accept. In the game, there are two active players - the investor and a partner. The investor is given a number of tokens to which a monetary value is assigned. They are then told they can give any number of tokens (including none) to the partner. The tokens they give will be multiplied by some factor, typically three or four times, and they get to keep any they did not give away. The partner then decides how many of these tokens they wish to return to the investor. The investor and partner are not allowed to communicate during the game. 

The level of trust between the two parties influences the amount of tokens transferred. If the partners are trusting and collaborate well, both players end up with more money than they started with. However, there is the opportunity for the partner to abuse trust; this is a risk to which the investor exposes themselves by transferring money in the first place.  It has been shown that participants in the Trust Game are more likely to exhibit risky financial behaviour when playing with a partner with a more trustworthy face \cite{van2014robot}.

\section{The Experiment}  \label{sec:EXP}

\subsection{The Setup}

The Trust Game outlined above was implemented as part of a custom Furhat `skill' in which parameters relating to eye and eyebrow shapes were adjusted to create potentially `trustworthy' and `untrustworthy' faces. The base face mesh chosen was the default Furhat `Alex'.  The instructions for the game were presented in written form so that there was no difference in their delivery.  The participants were not permitted to talk to the robot during the Game, e.g.\ to ask about the robot's intentions.  Participants were asked to play the Game with a series of robot partners exhibiting different faces, and the amount of money risked was recorded.

The experiment was conducted as a within-subjects factorial design. The independent variables were eye size, eye shape and eyebrow width, with each factor having two levels:
\begin{itemize}
\item Eye Size: Small or Large
\item Eye Shape: Narrow or Round
\item Eyebrow Width: Thin or Thick
\end{itemize}

\begin{figure}[h!]
      \centering
      \framebox{\parbox{\columnwidth}{
      \includegraphics[width=\columnwidth]{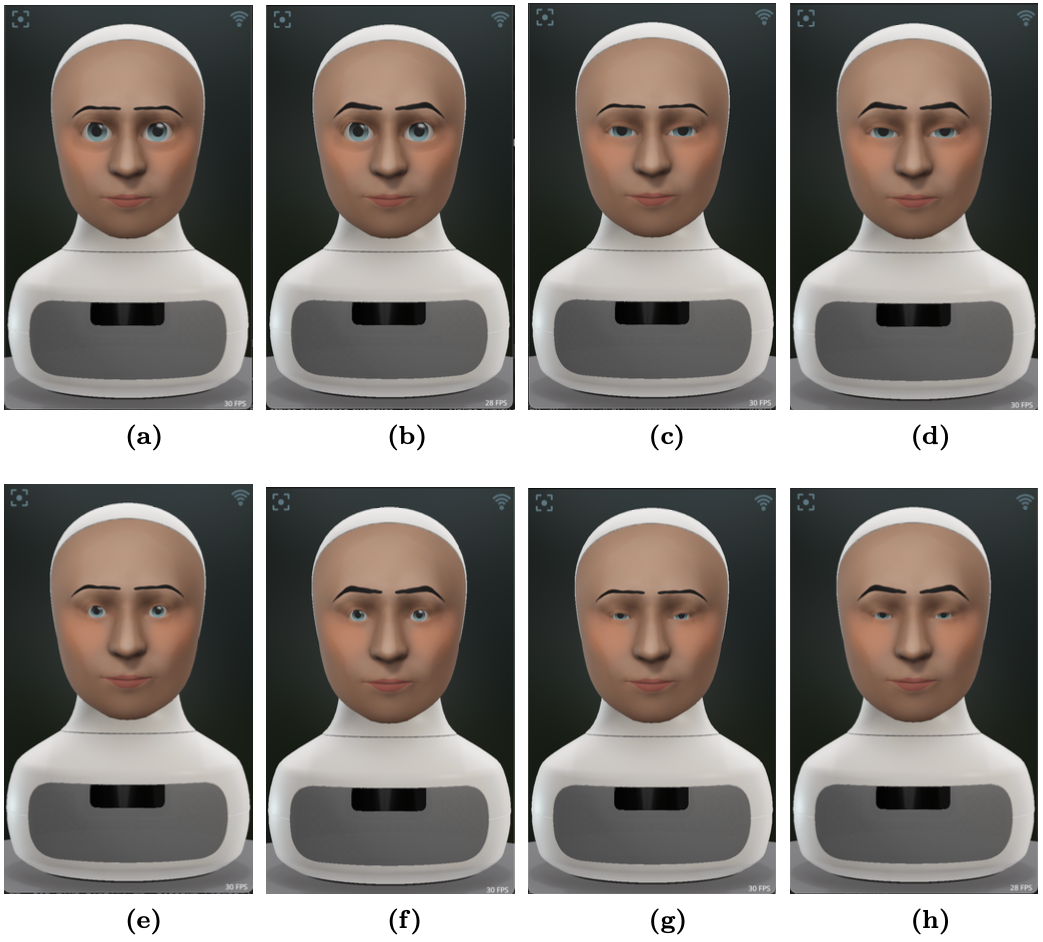}
      }}
      \caption{The eight robot faces used in this experiment (see Table~\ref{tab:TM} for details).  Note that the images are taken from the Furhat SDK, whereas the subjects were exposed to the actual robot.}
      \label{fig:FACES}
\end{figure}

This resulted in a 2x2x2 factorial design with a total of eight different robot face designs -- see Fig.~\ref{fig:FACES}.

\subsection{Participants}

A total of 32 participants were recruited from the University of Sheffield, with ages ranged from 19 to 50, with an average age of 25. Ten participants identified as female, 21 as male and one as non-binary. Participants were offered the chance to win one of two £25 vouchers for an online retailer to provide a motivation for both participation and playing the Trust Game seriously.

An a-priori power analysis carried out using G*Power software showed that, for the study to have 95\% power to identify a medium-size effect (using Cohen's f value of 0.25) with a significance level of 0.05, the minimum sample size required for statistical significance was 23. Hence, the final sample size of n=32 satisfied these requirements.

\subsection{Measures}

Trustworthiness was evaluated using two different methods. The first was a series of self-report measures relating to trust which were modified from the `Checklist for Trust between People and Automation' scale, a well-established twelve-item, seven-point Likert scale commonly used to evaluate trust between people and automation \cite{jian2000foundations}. The items were modified to replace the word ``\emph{system}'' with the word ``\emph{robot}'' to fit the context of the experiment. Some items were not included (e.g.\ ``\emph{The system provides security}''), as they were not relevant for the robot.  Others were removed in the interest of reducing redundancy and mitigating participant fatigue.  This resulted in a seven-point Likert scale for the following five statements:
\begin{itemize}
\item \emph{The robot is deceptive.}
\item \emph{I am suspicious of the robot's intent, action, or outputs.}
\item \emph{The robot has integrity.}
\item \emph{I can trust the robot.}
\item \emph{The robot is reliable.}
\end{itemize}

The second measure of perceived trustworthiness was via the economic Trust Game. Participants were asked to select how many tokens to give to each robot. They were not informed how each robot responded to their initial investment, as this would have resulted in undesirable order effects - the results of the previous robot were likely to have an effect on future investments. The independence of cases was an absolutely critical assumption of the Analysis of Variance (ANOVA) method that was used to analyse the data.

\subsection{Procedure}

On the day of the experiment, participants were invited to the HRI lab at the University of Sheffield where they sat at a desk facing the Furhat robot. They were presented with a participant information sheet and a consent form which they were required to read and complete before proceeding.

Each robot was rated using sliders in a web app, and a text field was provided for entering the amount to invest in the Trust Game. When a participant pressed \texttt{Next}, the Furhat projector was dimmed, the sliders were reset to default values and a timed message appeared in the web app informing them that the robot face would change. After three seconds, the projector was turned back on showing a different face. This was repeated for all eight robots, at which point the experiment concluded.

\section{Results} \label{sec:RES}

A factor analysis of the Likert-scale responses revealed that the data was unidimensional.  Hence, it was acceptable to calculate a composite score for trustworthiness by taking the means and variances of the responses to all five statements (introduced above).  The results are shown in Table~\ref{tab:TM}, and it can be seen that Robot (e), which has small round eyes and thin eyebrows (see Fig.~\ref{fig:FACES}), was perceived to be the most trustworthy. The least trustworthy robot was Robot (h), which has small narrow eyes and thick eyebrows. These findings are broadly in line with the hypothesis, with the exception that Robot (e) and Robot (f) were rated more positively than Robot (a). All four robots with big eyes tended toward variance, suggesting that eye size might be a polarising factor.

\begin{table}[!ht]
    \caption{Overall Trustworthiness Opinion Responses}\label{tab:TM}
    \centering
    \begin{tabular}{| l | c | c | c | c | c |}
    \hline
        \textbf{Condition} & \textbf{Eye Size} & \textbf{Eye Shape} & \textbf{Brow Width} & \textbf{$\mu$} & \textbf{$\sigma^2$}\\ \hline
        \hline
        Robot (a) & big & round & thin & 4.28 & 1.16 \\ \hline
        Robot (b) & big & round & thick & 4.02 & 1.12 \\ \hline
        Robot (c) & big & narrow & thin & 4.01 & 1.22 \\ \hline
        Robot (d) & big & narrow & thick & 3.87 & 1.22 \\ \hline
        Robot (e) & small & round & thin & 4.55 & 1.06 \\ \hline
        Robot (f) & small & round & thick & 4.38 & 0.91 \\ \hline
        Robot (g) & small & narrow & thin & 3.78 & 1.23 \\ \hline
        Robot (h) & small & narrow & thick & 3.33 & 1.01 \\ \hline
    \end{tabular}
\end{table}

A three-way analysis of variance (ANOVA) was conducted between the three fixed factors: eye size, eye shape and eyebrow width, and the composite trustworthiness score. These results are summarised in Table~\ref{tab:AT}.  It can be seen that significant results were found for Eye Shape and for Eye Size + Eye Shape.

\begin{table}[!ht]
    \caption{ANOVA Results for Trustworthiness Opinions}\label{tab:AT}
    \centering
    \begin{tabular}{| l | c | c |}
    \hline
        \textbf{Source} & \textbf{F} & \textbf{p} \\ \hline
        \hline
        Eye Size & 0.074 & 0.786 \\ \hline
        Eye Shape *** & 18.066 & 0.001 \\ \hline
        Brow Width & 3.724 & 0.055 \\ \hline
        Eye Size + Eye Shape ** & 7.096 & 0.008 \\ \hline
        Eye Size + Brow Width & 0.192 & 0.662 \\ \hline
        Eye Shape + Brow Width & 0.088 & 0.767 \\ \hline
        Eye Size + Eye Shape + Brow Width & 0.556 & 0.456 \\ \hline
    \end{tabular}
\end{table}

The same three-way ANOVA was carried out on the Trust Game responses, and the results are summarised in Table~\ref{tab:AG}. Again, a significant result was found for Eye Shape.

\begin{table}[h]
    \caption{ANOVA Results for the Trust Game}\label{tab:AG}
    \centering
    \begin{tabular}{| l | c | c |}
    \hline
        \textbf{Source} & \textbf{F} & \textbf{p} \\ \hline
        \hline
        Eye Size & 0.589 & 0.443 \\ \hline
        Eye Shape *** & 10.756 & 0.001 \\ \hline
        Brow Width & 1.520 & 0.219 \\ \hline
        Eye Size + Eye Shape & 0.156 & 0.693 \\ \hline
        Eye Size + Brow Width & 0.589 & 0.443 \\ \hline
        Eye Shape + Brow Width & 0.156 & 0.693 \\ \hline
        Eye Size + Eye Shape + Brow Width & 0.005 & 0.944 \\ \hline
    \end{tabular}
\end{table}

\subsubsection*{Eye Shape}
The analysis supported the hypothesis that round eyes are perceived as more trustworthy than narrow eyes. This was shown using both self-report measures and then confirmed by the results of the Trust Game. This finding is in line with research that suggests that eye area increases perceptions of attractiveness in humans, which is linked to the perception of honesty and trustworthiness \cite{zebrowitz1991impact,zebrowitz1993they}. 

\subsubsection*{Eyebrow Width}
There was a tendency to perceive robot faces with thin eyebrows as more trustworthy than thick eyebrows although the ANOVA results did not meet the formal requirements to be considered as statistically significant. 

\subsubsection*{Eye Size}
The hypothesis that trustworthiness increases with eye size was \underline{not} supported.  This was a surprising result given that many studies have confirmed this to be the case \cite{song2021effect,zebrowitz1991impact}. 

\subsubsection*{Interaction Effect: Eye Size + Eye Shape}
The only significant interaction effect found in this study was between eye size and eye shape. When eyes are small and narrow they are much less trusted than large, narrow eyes. This is presumably because the small narrow eyes are barely visible. However, this effect is reversed for round eyes; when the eyes are small and round they are more trusted than eyes which are large and round. A possible explanation for this is that the small round eyes are more proportionate to the facial structure than the large round ones (which, as implemented, looked unnatural -- discussed below).

\section{Discussion and Conclusion} \label{sec:CONC}

The study encountered some practical limitations due to the nature of Furhat's mask moulding.  In particular, it was found that adjusting the positions of facial features can lead to unnatural renderings.  For this reason, it was decided \emph{not} to move features around, but rather to adjust feature size and form only.  However, despite these mitigations, unstructured interviews with participants revealed the large eyes were perceived as ``\emph{very intense}'' or ``\emph{alien in appearance}'', suggesting that the manipulation of the face mesh may have been too extreme and the eyes had been made disproportionately large. Likewise, the deformation of the eyebrows in the particular face mesh were somewhat angular and not particularly realistic or representative of a typical eyebrow, especially that of a baby.

It should also be noted that eye shapes and sizes vary across different populations, so there is a possibility that this result might not be reproduced in another country with different participants. It is possible that participants are more likely to trust in robots who are designed with similar features to them, which would be an interesting avenue to explore in future work.

Despite these issues, the study reported here has shown that the design of facial features has a significant impact on the impression of trustworthiness on first encounters, in particular, the eyes. This initial positive impression is hugely important in certain fields, such as medical and therapeutic settings. Social robots are here to stay and it is of great importance to society that there is discussion surrounding how we would like our relationship with these novel agents to be should they one day be able to autonomously navigate our complex world. The appearance of robots is an enormous factor in this discussion, and a careful understanding of HRI and social needs is imperative for the resulting design and ultimately their acceptance.

\addtolength{\textheight}{-6cm}   

\bibliographystyle{IEEEtran}
\bibliography{IEEEabrv,mybibfile}

\begin{thebibliography}{10}
\providecommand{\url}[1]{#1}
\csname url@samestyle\endcsname
\providecommand{\newblock}{\relax}
\providecommand{\bibinfo}[2]{#2}
\providecommand{\BIBentrySTDinterwordspacing}{\spaceskip=0pt\relax}
\providecommand{\BIBentryALTinterwordstretchfactor}{4}
\providecommand{\BIBentryALTinterwordspacing}{\spaceskip=\fontdimen2\font plus
\BIBentryALTinterwordstretchfactor\fontdimen3\font minus
  \fontdimen4\font\relax}
\providecommand{\BIBforeignlanguage}[2]{{%
\expandafter\ifx\csname l@#1\endcsname\relax
\typeout{** WARNING: IEEEtran.bst: No hyphenation pattern has been}%
\typeout{** loaded for the language `#1'. Using the pattern for}%
\typeout{** the default language instead.}%
\else
\language=\csname l@#1\endcsname
\fi
#2}}
\providecommand{\BIBdecl}{\relax}
\BIBdecl

\bibitem{Deutsch1958}
M.~Deutsch, ``{Trust and suspicion},'' \emph{Journal of Conflict Resolution},
  vol.~2, no.~4, pp. 265--279, dec 1958.

\bibitem{Cominelli2021}
L.~Cominelli, F.~Feri, R.~Garofalo, C.~Giannetti, M.~A.
  Mel{\'{e}}ndez-Jim{\'{e}}nez, A.~Greco, M.~Nardelli, E.~P. Scilingo, and
  O.~Kirchkamp, ``{Promises and trust in human–robot interaction},''
  \emph{Scientific Reports}, vol.~11, no.~1, p. 9687, may 2021.

\bibitem{Baier2014}
A.~Baier, \emph{{Trust and antitrust}}.\hskip 1em plus 0.5em minus 0.4em\relax
  New York, NY, USA: Routledge, 2014, ch.~31, p.~26.

\bibitem{Sanders2011}
T.~Sanders, K.~E. Oleson, D.~R. Billings, J.~Y.~C. Chen, and P.~A. Hancock,
  ``{A model of human-robot trust: Theoretical model development},''
  \emph{Proceedings of the Human Factors and Ergonomics Society Annual
  Meeting}, vol.~55, no.~1, pp. 1432--1436, sep 2011.

\bibitem{Kwon2016}
M.~Kwon, M.~F. Jung, and R.~A. Knepper, ``{Human expectations of social
  robots},'' in \emph{2016 11th ACM/IEEE International Conference on
  Human-Robot Interaction (HRI)}.\hskip 1em plus 0.5em minus 0.4em\relax
  Christchurch, New Zealand: IEEE, mar 2016, pp. 463--464.

\bibitem{Sanders2017}
T.~L. Sanders, K.~MacArthur, W.~Volante, G.~Hancock, T.~MacGillivray,
  W.~Shugars, and P.~A. Hancock, ``{Trust and prior experience in human-robot
  interaction},'' \emph{Proceedings of the Human Factors and Ergonomics Society
  Annual Meeting}, vol.~61, no.~1, pp. 1809--1813, sep 2017.

\bibitem{Robert2018}
L.~P. Robert, ``{Personality in the human robot interaction literature: A
  review and brief critique},'' in \emph{Proceedings of the 24th Americas
  Conference on Information Systems}.\hskip 1em plus 0.5em minus 0.4em\relax
  New Orleans, LA, USA: SSRN, 2018.

\bibitem{Huang2023d}
G.~Huang and R.~K. Moore, ``{Get off on the right foot with whom?: How users'
  profiles affect their perception and experience with a social robot},'' in
  \emph{IEEE International Conference on Robotics and Automation (ICRA)},
  London, 2023.

\bibitem{PARK2008}
E.~Park, Q.~Jenkins, and X.~Jiang, ``{Measuring trust of human operators in new
  generation rescue robots},'' \emph{Proceedings of the JFPS International
  Symposium on Fluid Power}, no. 7-2, pp. 489--492, 2008.

\bibitem{Wang2010}
L.~Wang, P.-L.~P. Rau, V.~Evers, B.~K. Robinson, and P.~Hinds, ``{When in Rome:
  The role of culture and context in adherence to robot recommendations},'' in
  \emph{2010 5th ACM/IEEE International Conference on Human-Robot Interaction
  (HRI)}.\hskip 1em plus 0.5em minus 0.4em\relax Osaka, Japan: ACM Press, 2010,
  pp. 359--366.

\bibitem{Goetz2003}
J.~Goetz, S.~Kiesler, and A.~Powers, ``{Matching robot appearance and behavior
  to tasks to improve human-robot cooperation},'' in \emph{The 12th IEEE
  International Workshop on Robot and Human Interactive Communication, 2003.
  Proceedings. ROMAN 2003.}\hskip 1em plus 0.5em minus 0.4em\relax Millbrae,
  CA, USA: IEEE, 2003, pp. 55--60.

\bibitem{Hancock2011}
P.~A. Hancock, D.~R. Billings, K.~E. Schaefer, J.~Y.~C. Chen, E.~J. de~Visser,
  and R.~Parasuraman, ``{A meta-analysis of factors affecting trust in
  human-robot interaction},'' \emph{Human Factors: The Journal of the Human
  Factors and Ergonomics Society}, vol.~53, no.~5, pp. 517--527, oct 2011.

\bibitem{Eyssel2012}
F.~Eyssel, D.~Kuchenbrandt, S.~Bobinger, L.~de~Ruiter, and F.~Hegel, ``{'If you
  sound like me, you must be more human'},'' in \emph{Proceedings of the
  seventh annual ACM/IEEE international conference on Human-Robot
  Interaction}.\hskip 1em plus 0.5em minus 0.4em\relax New York, NY, USA: ACM,
  mar 2012, pp. 125--126.

\bibitem{Mori1970}
M.~Mori, ``{Bukimi no tani (the uncanny valley)},'' \emph{Energy}, vol.~7, pp.
  33--35, 1970.

\bibitem{Moore2012}
R.~K. Moore, ``{A Bayesian explanation of the ‘Uncanny Valley' effect and
  related psychological phenomena},'' \emph{Nature Scientific Reports}, vol.~2,
  no. 864, 2012.

\bibitem{willis2006first}
J.~Willis and A.~Todorov, ``{First impressions: Making up your mind after a
  100-ms exposure to a face},'' \emph{Psychological Science}, vol.~17, no.~7,
  pp. 592--598, jul 2006.

\bibitem{van2008friend}
M.~{van 't Wout} and A.~Sanfey, ``{Friend or foe: The effect of implicit
  trustworthiness judgments in social decision-making},'' \emph{Cognition},
  vol. 108, no.~3, pp. 796--803, sep 2008.

\bibitem{berry1988s}
D.~S. Berry and L.~Zebrowitz-McArthur, ``{What's in a face? Facial maturity and
  the attribution of legal responsibility},'' \emph{Personality and Social
  Psychology Bulletin}, vol.~14, no.~1, pp. 23--33, mar 1988.

\bibitem{zebrowitz1991impact}
L.~A. Zebrowitz and S.~M. McDonald, ``{The impact of litigants' baby-facedness
  and attractiveness on adjudications in small claims courts.}'' \emph{Law and
  Human Behavior}, vol.~15, no.~6, pp. 603--623, 1991.

\bibitem{jaeger2020can}
B.~Jaeger, A.~T. Todorov, A.~M. Evans, and I.~van Beest, ``{Can we reduce
  facial biases? Persistent effects of facial trustworthiness on sentencing
  decisions},'' \emph{Journal of Experimental Social Psychology}, vol.~90, p.
  104004, sep 2020.

\bibitem{kalegina2018characterizing}
A.~Kalegina, G.~Schroeder, A.~Allchin, K.~Berlin, and M.~Cakmak,
  ``{Characterizing the design space of rendered robot faces},'' in
  \emph{Proceedings of the 2018 ACM/IEEE International Conference on
  Human-Robot Interaction}.\hskip 1em plus 0.5em minus 0.4em\relax New York,
  NY, USA: ACM, feb 2018, pp. 96--104.

\bibitem{song2021effect}
Y.~Song, A.~Luximon, and Y.~Luximon, ``{The effect of facial features on facial
  anthropomorphic trustworthiness in social robots},'' \emph{Applied
  Ergonomics}, vol.~94, p. 103420, jul 2021.

\bibitem{Ghazali2018}
A.~S. Ghazali, J.~Ham, E.~I. Barakova, and P.~Markopoulos, ``{Effects of robot
  facial characteristics and gender in persuasive human-robot interaction},''
  \emph{Frontiers in Robotics and AI}, vol.~5, jun 2018.

\bibitem{desteno2012detecting}
D.~DeSteno, C.~Breazeal, R.~H. Frank, D.~Pizarro, J.~Baumann, L.~Dickens, and
  J.~J. Lee, ``{Detecting the trustworthiness of novel partners in economic
  exchange},'' \emph{Psychological science}, vol.~23, no.~12, pp. 1549--1556,
  2012.

\bibitem{zebrowitz1992impressions}
L.~A. Zebrowitz and J.~M. Montepare, ``{Impressions of babyfaced individuals
  across the life span.}'' \emph{Developmental Psychology}, vol.~28, no.~6, pp.
  1143--1152, nov 1992.

\bibitem{chita2021can}
M.~Chita-Tegmark, T.~Law, N.~Rabb, and M.~Scheutz, ``{Can you trust your trust
  measure?}'' in \emph{Proceedings of the 2021 ACM/IEEE International
  Conference on Human-Robot Interaction}.\hskip 1em plus 0.5em minus
  0.4em\relax New York, NY, USA: ACM, mar 2021, pp. 92--100.

\bibitem{berg1995trust}
J.~Berg, J.~Dickhaut, and K.~McCabe, ``{Trust, reciprocity, and social
  history},'' \emph{Games and Economic Behavior}, vol.~10, no.~1, pp. 122--142,
  jul 1995.

\bibitem{van2014robot}
R.~van~den Brule, R.~Dotsch, G.~Bijlstra, D.~H.~J. Wigboldus, and P.~Haselager,
  ``{Do robot performance and behavioral style affect human trust?}''
  \emph{International Journal of Social Robotics}, vol.~6, no.~4, pp. 519--531,
  nov 2014.

\bibitem{jian2000foundations}
J.-Y. Jian, A.~M. Bisantz, and C.~G. Drury, ``{Foundations for an empirically
  determined scale of trust in automated systems},'' \emph{International
  Journal of Cognitive Ergonomics}, vol.~4, no.~1, pp. 53--71, mar 2000.

\bibitem{zebrowitz1993they}
L.~A. Zebrowitz, J.~M. Montepare, and H.~K. Lee, ``They don't all look alike:
  Individual impressions of other racial groups.'' \emph{Journal of personality
  and social psychology}, vol.~65, no.~1, p.~85, 1993.

\end{thebibliography}

\end{document}